\documentclass[runningheads]{llncs}

\usepackage{eccv}

\usepackage{eccvabbrv}
\usepackage{verbatim}
\usepackage{graphicx}
\usepackage{booktabs}

\usepackage[accsupp]{axessibility}

\usepackage{orcidlink}
\usepackage{multirow}
\usepackage{pifont}
\usepackage{algorithm}
\usepackage{algpseudocode}
\usepackage{graphicx}
\usepackage{makecell}
\usepackage{soul}
\usepackage{microtype}
\usepackage{wrapfig}

\def\SmD{SmD}
\def\CrD{CrD}
\def\ModelName{STAT}
\def\ModelFullName{Self-supervised Temporal Adaptive Teacher}

\begin{document}

\title{\ModelName: Towards \\ Generalizable Temporal Action Localization}

\author{Yangcen Liu\inst{1} 
\and
Ziyi Liu\inst{2}
\and
Yuanhao Zhai\inst{3}
\and
Wen Li\inst{1}
\and
David Doerman\inst{3}
\and
Junsong Yuan\inst{3}
}

\authorrunning{Liu et al.}

\institute{University of Electronic Science and Technology of China, China \and
University of Science and Technology Beijing, China\\
\and
University at Buffalo, USA}

\maketitle

\begin{abstract}
Weakly-supervised temporal action localization (WTAL) aims to recognize and localize action instances with only video-level labels.
Despite the significant progress, existing methods suffer from severe performance degradation when transferring to different distributions and thus may hardly adapt to real-world scenarios.
To address this problem, we propose the Generalizable Temporal Action Localization task (GTAL), which focuses on improving the generalizability of action localization methods.
We observed that the performance decline can be primarily attributed to the lack of generalizability to different action scales.
To address this problem, we propose \ModelName~(\ModelFullName), which leverages a teacher-student structure for iterative refinement. Our \ModelName~features a refinement module and an alignment module.
The former iteratively refines the model's output by leveraging contextual information and helps adapt to the target scale.
The latter improves the refinement process by promoting a consensus between student and teacher models.
We conduct extensive experiments on three datasets, THUMOS14, ActivityNet1.2, and HACS, and the results show that our method significantly improves the baseline methods under the cross-distribution evaluation setting, even approaching the same-distribution evaluation performance.
\end{abstract}    
\section{Introduction}
\label{sec:intro}

Weakly-supervised Temporal Action Localization (WTAL), which focuses on using video-level annotations to identify and classify actions in time, has attracted considerable attention~\cite{autoloc,acmnet,co2net,cascade,actionunit,weaklysuperviseduncertainty,adaptivetwostream}.
Despite the advancement, most existing methods operate under the assumption that training and testing data are independent and identically distributed, but this assumption often does not hold in real-world scenarios.
Consequently, their practical applications greatly suffer due to poor generalization ability across different data distributions.

Although there exist domain adaptation (DA) methods to address domain shift issues \cite{segscale, odda}, applying them to Temporal Action Localization (TAL) presents significant challenges.
Firstly, acquiring video data and annotations is often difficult, making it challenging to train on combinations of different datasets or perform domain adaptation between them, thus limiting their applicability in TAL.
Secondly, untrimmed video data for TAL inherently contains various gaps, such as visual, annotation, and montage gaps (e.g., from GIF to Film), which are more pronounced compared to the data gaps typically encountered in conventional DA settings. As a result, leveraging existing DA methods in TAL becomes inherently more complex.

\begin{figure}[t]
  \centering
  \includegraphics[width=0.9\linewidth]{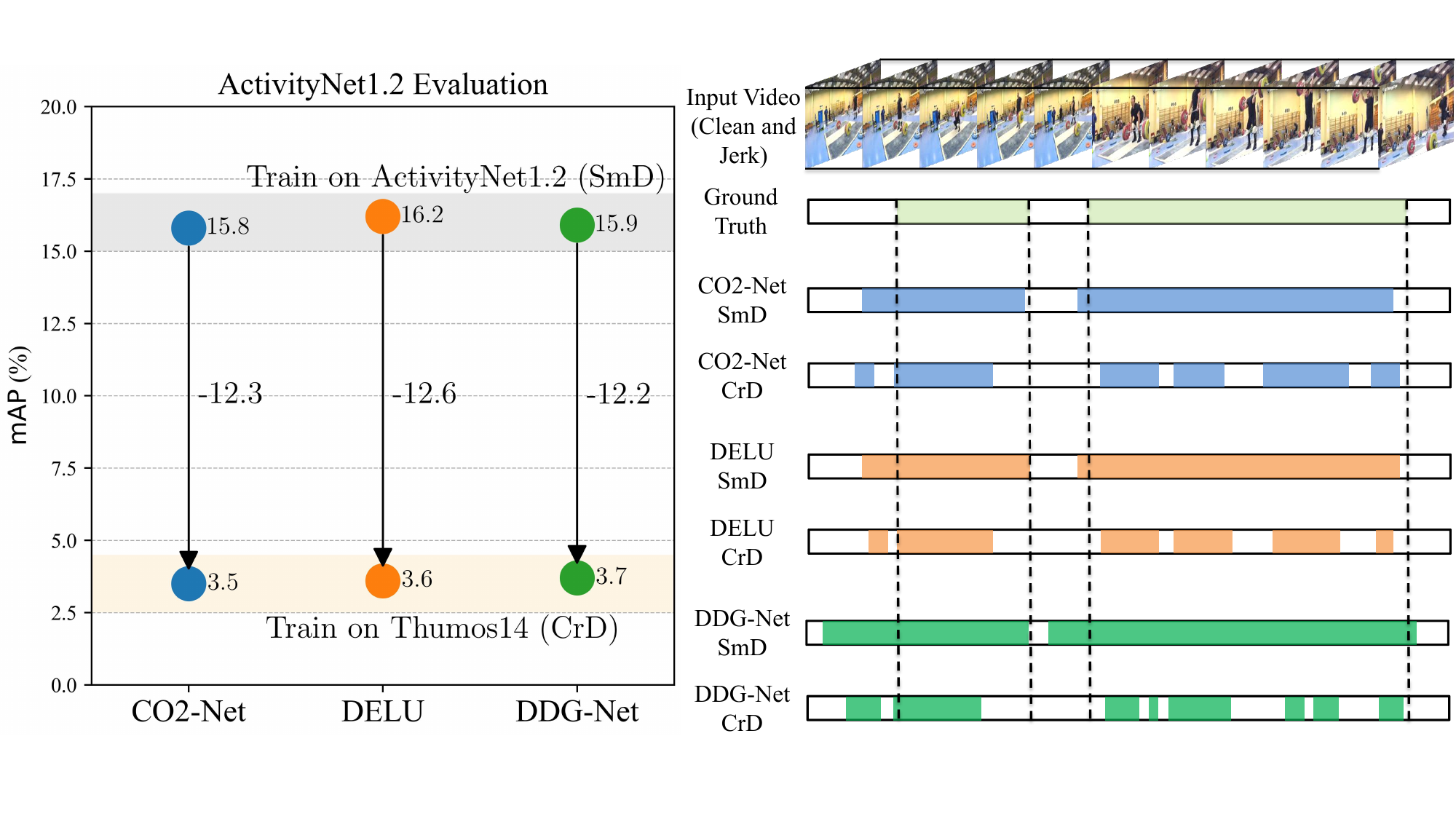}
  \caption{Performance comparison between same-distribution (SmD) and cross-distribution (CrD) evaluation. Left: Under the CrD setting\protect\footnotemark, state-of-the-art methods demonstrate significant performance degradation. Right: Compared to the SmD results, the CrD results appear to be more fragmented, highlighting a lack of adaptability in these methods to the temporal scale variations of the target distribution.}
  \label{oodgap}
\end{figure}

To address this problem, we introduce a novel setting, termed Generalizable Temporal Action Localization (GTAL).
Specifically, GTAL consists of two settings: training and evaluating on the \emph{sharing action categories} of the \textbf{s}a\textbf{m}e-\textbf{d}istribution (SmD), and \textbf{cr}oss-\textbf{d}istribution evaluation (CrD).
The SmD setting overlaps with the traditional evaluation protocol, while the CrD setting, which evaluates the generalization ability, is rarely used previously.
We found that state-of-the-art WTAL methods are vulnerable to handling the distribution shift, leading to significant performance degradation in the CrD setting, as evidenced in \cref{oodgap}.
Thus, there is a pressing need for analyzing the address the poor generalization ability of existing methods.

\footnotetext{We use the shared 8 classes from THUMOS14~\cite{thumos} and ActivityNet1.2~\cite{activitynet} to conduct the evaluation.}

We analyze the root causes of the CrD performance drop from the classification
and localization aspects in \cref{problem}.Our findings indicate that while most
methods maintain consistent action classification accuracy across SmD and CrD
settings, their \emph{localization performance} significantly deteriorates in
the CrD setting. This suggests that localization is the primary challenge rather
than classification in GTAL, and we attribute the degraded localization to the
scale variance and visual gap.

To enhance the model's generalization capabilities across various scales and
mitigate visual discrepancies, we propose \textbf{S}elf-supervised
\textbf{T}emporal \textbf{A}daptive \textbf{T}eacher (\ModelName). Our core
insight is that, despite the localization performance degradation, existing
methods maintain a high snippet-level classification accuracy on the snippets
with high activations (\cref{problem}). Building on this observation, our
\ModelName employs a teacher-student framework, wherein the teacher model,
characterized by its robust snippet-level classification accuracy, serves as a
guide for the student model. This guidance facilitates the learning process for
ambiguous snippets, and thus boosting the generalization ability of the student
model.

Besides the student-teacher structure, our \ModelName~features two designs: a refinement module and an alignment module.
The refinement module focuses on adapting the attention scale derived from the teacher model to the \CrD~dataset.
It employs a salience sampling strategy to gather contextual information and refine the central snippet's output.
This module also incorporates a rule to maintain ranking consistency in attention scores for the semantic relationship between relevant snippets.
Then, the alignment module focuses on mitigating the impact of changes in attention scores induced by the refinement module with different priors.
Also, with the weak augmentation of the input video feature of the student model, the module improves its robustness.
The alignment module independently aligns the output classification score and attention, directing the student model's training. Additionally, it calibrates the classification score with the attention to regulate the classification of foreground and background.
The integration of these modules in \ModelName~allows a pre-trained TAL model to be effectively adapted from a \SmD to a \CrD setting, resulting in comparatively high performance, significantly improving the baseline methods.

To summarize, our contribution is threefold:
\begin{itemize}
    \item We introduce the generalizable temporal action localization (GTAL) task, emphasizing the generalization ability across datasets. We further analyze the reasons causing the performance drop of existing methods in GTAL, and find that the main cause lies in the degraded localization.
    \item We propose \ModelName for GTAL, based on a teacher-student structure improving better robustness, which features two novel designs: refinement modules and an alignment module. The former iteratively refines the teacher model to the target dataset, and the latter aligns the teacher and student outputs, thus promoting precise \CrD localization.
    \item We conduct comprehensive experiments on three WTAL benchmarks: THUMOS14~\cite{thumos}, ActivityNet1.2~\cite{activitynet} and HACS~\cite{hacs}. Our
    \ModelName~enables existing methods to better adapt different data distribution, achieving significant \CrD performance improvement.
\end{itemize}
\section{Related Work}
\label{sec:related}

\textbf{Weakly-supervised Temporal Action Localization} has been widely studied~\cite{sparsepooling,actionunit,temporalsegment,hideandseek,stepbystep,autoloc}.
Recently, there exists a line a work~\cite{weaklyknoledge,pmil,highlight,refineloc,twostream} that attempts to generate pseudo labels to iteratively guide the model training.
P-MIL~\cite{pmil} uses a two-stage framework, using pseudo labels to guide the training.
Another series of work~\cite{twostream,adaptivetwostream,co2net,ddgnet,saliantwtal} exploit the relationship between cross-modal channels. CO2-Net~\cite{co2net} applies a cross-modal attention mechanism to enhance feature consistency within the structure. DDG-Net~\cite{ddgnet} directly designs a residual graph structure module to construct the cross-modal connection. Recently, a more efficient and popular method is to guide the training with uncertainty learning strategy~\cite{weaklysuperviseduncertainty,ugct,delu,cascade}. DELU~\cite{delu} introduces evidential learning~\cite{edlsurvey} to guide the sampling strategy and learning.
However, the effectiveness of the aforementioned strategies heavily relies on pre-set hyper-parameters and can overfit \SmD~datasets.
Our experiments demonstrate that those methods confront severe performance degradation under \CrD, rendering them unsuitable for GTAL.

\noindent \textbf{OOD Generalization} is proposed to improve the generalization ability of models~\cite{oodsurvey}.
As data from the target domains is not available, assumptions or prior knowledge about the data distributions are necessary for effective generalization.
Mainstream methods include data augmentation~\cite{dataaugmentsurvey}, leveraging generative adversarial losses~\cite{cyclegan}, causal inference~\cite{dacasual}, and risk minimization~\cite{riskmin,irm}.
Another line of approach such as~\cite{dginv,dgadv,dgcadv} addresses domain shift by seeking invariant distributions across domains.

\noindent \textbf{Scale Variation Handling} Scale variation is a phenomenon where models perform poorly over different scales.
In GTAL, we find that scale variation is the primary factor contributing to the performance degradation (\cref{problem}). Similar problems have been widely researched in different tasks, such as image classification~\cite{multiimage,dialated,crossvit,mvit}, semantic segmentation~\cite{multiscaleseg,multivitseg,hrda}, and object detection~\cite{fpn,snip,sniper,dst,autofocus}, \etc.
Specifically, the SNIP series of work~\cite{snip,sniper,autofocus} leverages image pyramids to normalize object scales during multi-scale training, thus to better detecting small objects.
Supplementary strategies like context modeling~\cite{contextsod} and data augmentation~\cite{sodaugmentation} are commonly adopted to enhance performance.
In WTAL, several methods leverage the feature pyramid network~\cite{fpn} for the scale variation problem~\cite{cascade,multiscale_wtal,convex}, but most of them focus on the SmD setting.
However, in GTAL, the absence of instance-level supervision for each action instance presents a significant challenge, making direct normalization of action instances to a specific scale a complex task.
\section{Challenges in GTAL}
\label{problem}

State-of-the-art methods are observed to experience significant performance degradations in the cross-distribution (\CrD) ~testing scenarios.
In this section, we analyze the main factors that affect the \CrD~performance from the \textbf{classification} and \textbf{localization} aspects, respectively.
Our finding suggests that the main factor that causes the \CrD~performance degradation lies in poor localization ability.

\subsection{Problem Setup}

\noindent\textbf{Generalizable Temporal Action Localization (GTAL)GTAL.}
Our generalizable temporal action localization (GTAL) consists of two experimental settings: same-distribution (\SmD) for training and testing on different splits of the same datasets; and cross-distribution (\CrD) for using different datasets for training and testing, respectively.
Note that the evaluation is only conducted on the shared action classes among the different datasets.

\noindent\textbf{Baseline.} 
To analyze the GTAL task, we choose DELU~\cite{delu} as the baseline.
We begin by describing a general baseline ~\cite{co2net,delu,ddgnet}.
Following recent methods~\cite{untrimmednets, pmil}, we use off-the-shelf RGB and optical flow snippet-level features as input, denoted as $\boldsymbol{F} \in \mathbb{R}^{N \times D}$, where $N$ indicates the number of snippets.
Given the snippet-level features, a category-agnostic attention branch $\mathcal{M}^a$ is used to get the attention value $\boldsymbol{\varphi} \in \mathbb{R} ^{T\times 1}$, and a classification branch is used to predict the Class Activation Sequence (CAS) $\boldsymbol{\Psi} \in \mathbb{R}^{T\times (C_I + 1)}$, where $C_I + 1$ indicates the number of shared classes plus the background class.
After that, based on the assumption that background is present in all videos but filtered out by the attention sequence $\boldsymbol{\varphi}$, the original predicted video-level classification scores $\boldsymbol{\hat{y}}_{base} \in \mathbb{R}^{C_{I}+1}$, and the foreground video-level classification scores $\boldsymbol{\hat{y}}_{supp} \in \mathbb{R}^{C_I+1}$ suppressed by attention, are derived by applying a temporal top-k aggregation strategy $f_{agg}$~\cite{mil,untrimmednets}, respectively:
\begin{equation}
    \boldsymbol{\hat{y}}_{base} = f_{agg}(\boldsymbol{\Psi}),  \;
    \boldsymbol{\hat{y}}_{supp} = f_{agg}(\boldsymbol{\varphi} \odot \boldsymbol{\Psi}),
\label{equ:base_sup}
\end{equation}
where $\odot(\cdot)$ is the element wise production.

To consider background information of each video, we extend the video-level label: $\boldsymbol{y}_{base} = [\boldsymbol{y}, 1] \in \mathbb{R}^{C_I+1}$ and $\boldsymbol{y}_{supp} = [\boldsymbol{y}, 0] \in \mathbb{R}^{C_I+1}$. Guided by the video-level category label $\boldsymbol{y} \in \mathbb{R}^{C_I}$, the classification loss is defined as:
\begin{equation}
L_{cls} = -\sum_{c=1}^{C_I+1}\boldsymbol{y}_{base}\log \hat{\boldsymbol{y}}_{base} + \boldsymbol{y}_{supp}\log \boldsymbol{\hat{y}}_{supp}.
\label{equ:label}
\end{equation}
After training, the final localization results are obtained by thresholding the attention sequence.

\subsection{Analysis on Classification and Localization}
\label{problemlabel}

\begin{figure}[t]
  \centering
  \begin{subfigure}[b]{0.4\linewidth}
    \centering
    \includegraphics[width=\linewidth]{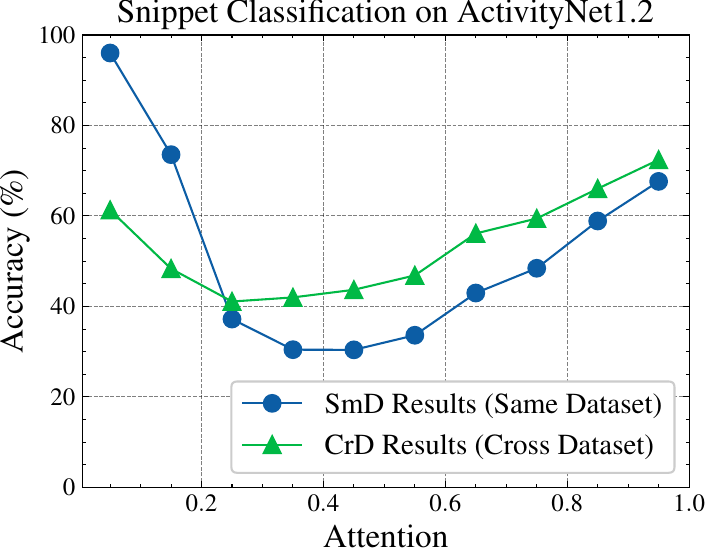}
    \caption{Classification accuracy of snippets with different attentions under \SmD and \CrD settings. Snippets with high or low attention tend to outperform those with ambiguous attention.}
    \label{snippet-classification-graph}
  \end{subfigure}
  \hfill
  \begin{subfigure}[b]{0.45\linewidth}
    \centering
    \includegraphics[width=\linewidth]{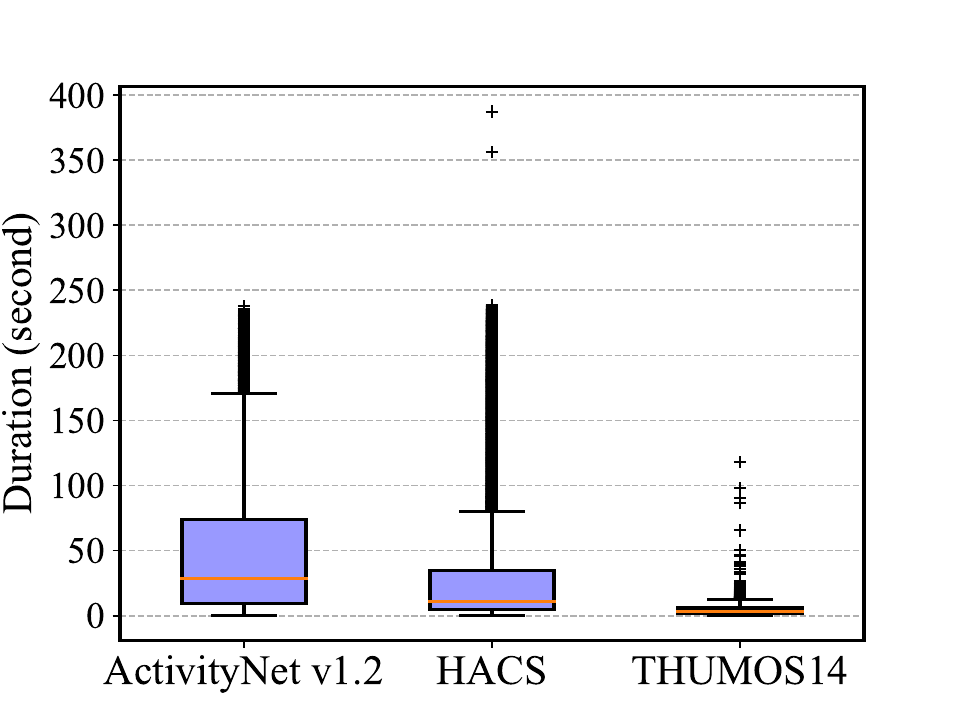}
    \caption{Action instance duration distribution in three datasets. The median duration in THUMOS14~\cite{thumos} is 3.0 seconds, while those in ActivityNet1.2~\cite{activitynet} and HACS~\cite{hacs} are 28.5 seconds and 11.2 seconds, respectively.}
    \label{duration}
  \end{subfigure}
  \caption{Snippet classification results and dataset statistics.}
  \label{combined-figure}
\end{figure}

\begin{wraptable}{r}{5.8cm}
    \resizebox{\linewidth}{!}{
    \begin{tabular}{c|>{\centering\arraybackslash}p{1.8cm}|>{\centering\arraybackslash}p{2.5cm}}
        \toprule
        Setting & SmD & CrD \\
        \midrule
        Video-level cls. acc. ($\mathcal{A}_v$) & $94.2\%$ & $91.3\%$ ($\color{blue}-2.9\%$) \\
        \hline
        Snippet-level cls. acc. ($\mathcal{A}_s$) & $51.2\%$& $42.0\%$ ($\color{blue} -9.2\%$) \\
        \hline
        \makecell{High-attention snippets \\ snippet-level cls. acc. ($\hat{\mathcal{A}}_s$)} & $84.7\%$  & $87.6\%$ ($\color{red}+2.9\%$)  \\
        \bottomrule
    \end{tabular}
    }
    \caption{The classification accuracy of DELU tested on ActivityNet1.2. High attention snippets are those with $\boldsymbol{\varphi}(n) >0.9$.}
    \label{classification-accuracy}
\end{wraptable}

We conduct a detailed analysis and a series of experiments using DELU. For the
\SmD setting, both training and evaluation are carried out on ActivityNet1.2.
For the \CrD setting, we train on THUMOS14 and evaluate on ActivityNet1.2.

\noindent \textbf{Classification analysis.} Transitions between distributions often lead to changes in semantics and appearance, potentially causing incorrect action classification and, consequently, a notable decline in \CrD localization
performance. 
To explore the possibility, we first conducted a video-level classification in \cref{classification-accuracy}.
Notably, both \SmD and \CrD video-level accuracy ($\mathcal{A}_v$) exceed 90\%, differing by a modest 2.9\%. Yet, as highlighted in \cref{oodgap}, there is a significant 12.4\% mean Average Precision (mAP) discrepancy between \SmD and \CrD
localization results.
This suggests that despite minor variations in visual and semantic across distributions, the model retains relatively high classification accuracy. 
Therefore, we infer that the primary contributor to the observed
performance degradation in CrD settings is not the classification accuracy.

\noindent \textbf{Localization analysis.}
To gain a clearer understanding of the factors influencing the GTAL performance, we further conduct a set of experiments on snippet-level classification, which is crucial for accurately detecting action proposal boundaries.
We present the results in two categories: (1) the overall classification accuracy of all snippets $\mathcal{A}_s$, considering only those snippets that contain shared action classes, and (2) the classification accuracy of high-attention snippets $\hat{\mathcal{A}}_s$, evaluated on snippets where the attention level exceeds 90\%.
These results are presented in \cref{classification-accuracy}.
Additionally, we performed a statistical analysis correlating $\mathcal{A}_s$ with attention levels in \cref{snippet-classification-graph}.

\begin{figure}[t]
    \centering
    \includegraphics[width=0.8\linewidth]{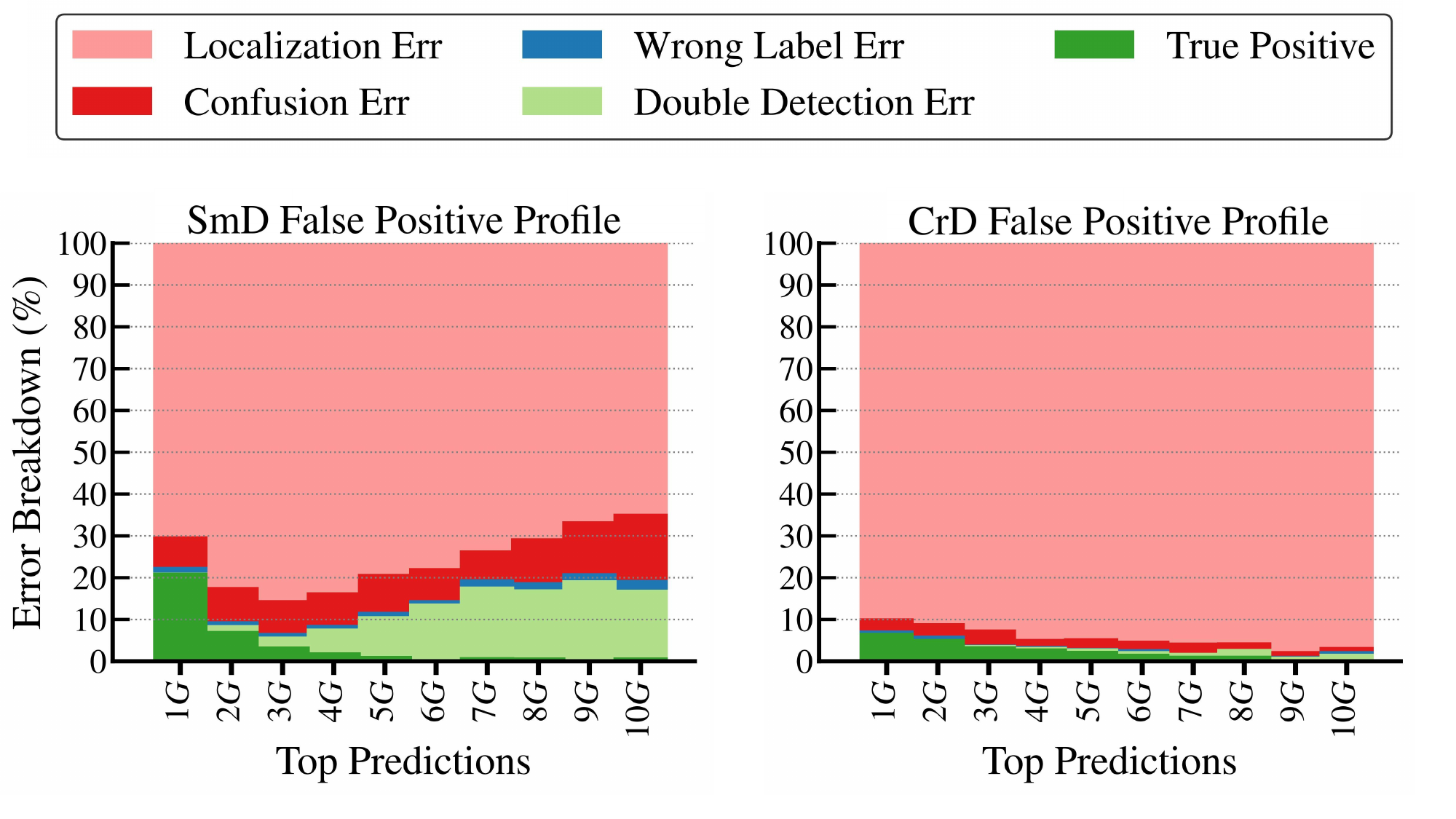}
    \caption{DETAD~\cite{diagnose} analysis of DELU prediction. Left: \SmD results. Right: \CrD results. Compared with \SmD, \CrD predictions contain much more localization error.}
    \label{fig:detad}
\end{figure}

\begin{figure*}[t]
  \centering
       \includegraphics[width=1.0\linewidth]{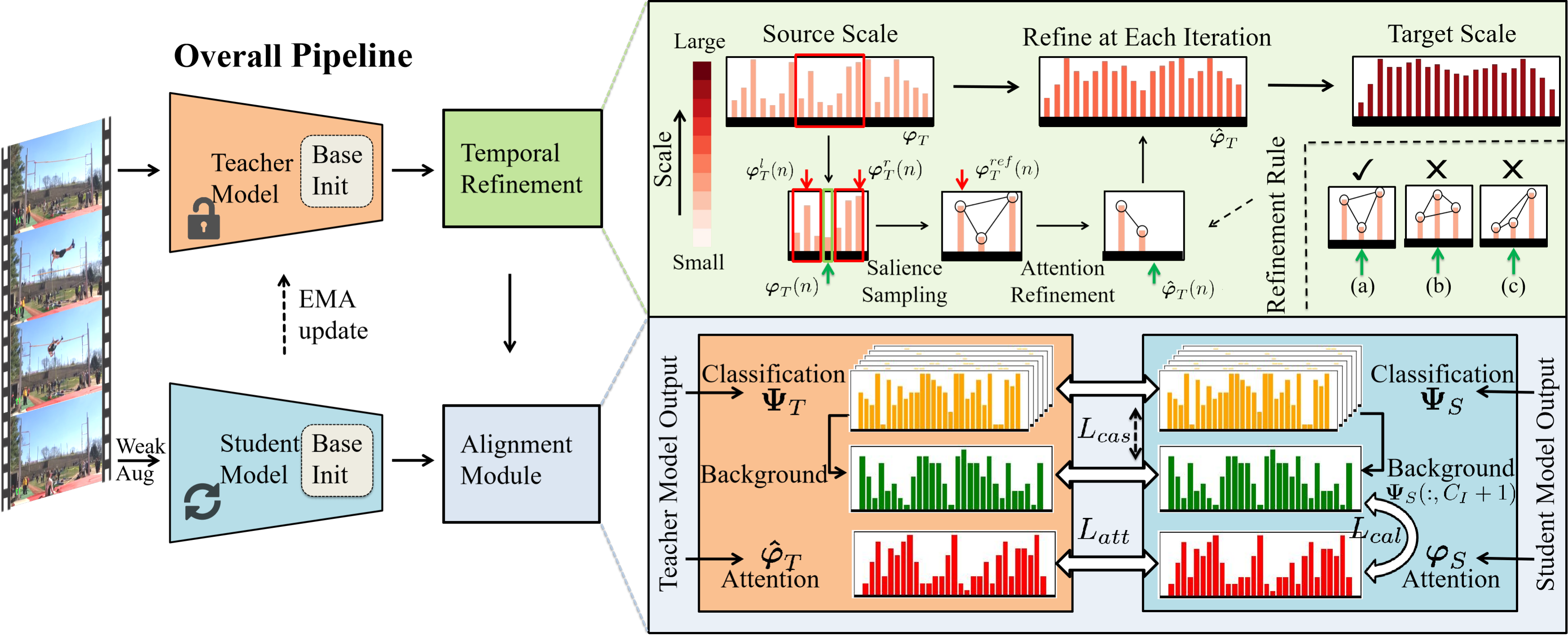}
       \caption{Overall framework of our \ModelFullName. The pipeline is built on the mean teacher framework. First, the teacher model and the student model are initialized with the \SmD model. Then, for each input video, the teacher model and the student model separately predict the CAS and attention. After that, the predicted attention from the teacher is refined in the temporal refinement module. Finally, the output of the student and teacher model is aligned in the alignment module, guiding the student to adapt to the target scale.}
       \label{pipeline}
\end{figure*}

The results show that, in comparison to video-level classification, the classification accuracy of high-attention snippets ($\hat{\mathcal{A}}_s$) in both SmD and CrD scenarios decreases by a modest 6.6\%.
In contrast, the classification accuracy of all snippets ($\mathcal{A}_s$) suffers from a significant drop, declining by 40.1\% in~\SmD and 51.3\% in~\CrD.
Furthermore, there is a noticeable pattern within the $\mathcal{A}_s$ data: snippets with high or low attention levels tend to maintain relatively high accuracy.
However, snippets that fall into the ambiguous middle range exhibit significantly lower classification accuracy (\cref{snippet-classification-graph}).
These intermediate attention snippets are crucial in determining action boundaries during the watershed thresholding inference process.
Thus, we propose that the misclassification of these snippets is a primary factor behind the poor CrD performance.

Video's defining characteristic of temporal continuity sets it apart from other modalities, which inspires us to hypothesize that the presence of ambiguous snippets in videos is linked to the duration of the action. Given that actions inherently possess a temporally continuous nature, factors like motion blur, video editing techniques, and subjects exiting the frame can lead to the emergence of these ambiguous snippets. Supporting this hypothesis, our findings reveal a notable variation in action durations across different datasets (\cref{duration}). The variance in duration is not just across datasets but is also observable within the same action category when comparing different datasets.

To facilitate our analysis, we further conduct DETAD analysis (\cref{fig:detad}) on the predictions, and the results reveal that the \CrD predictions contain more localization error than that in the \SmD setting\footnote{Unlike DETAD, we redefine localization error as instances overlap with the ground truth, but the intersection over union (IoU) does not meet the required threshold.}.

\noindent\textbf{Conclusion.}
According to the analysis above, we make two conclusions.
(1) The main reason for the performance degradation under the CrD setting is the temporal scale between action instances from different datasets.
(2) Snippets with high attention are salient and could maintain high classification accuracy in both SmD and CrD settings, which could be exploited to facilitate GTAL.
\section{Method}
\label{method}

The proposed \ModelFullName (\ModelName) is presented in~\cref{pipeline}. The
training pipeline consists of two stages. In the first stage, the base model is
trained similar to that of the traditional WTAL task. The second stage involves
adapting the model to a new distribution using a self-supervised pipeline. The
output of the base model for an input video comprises Attention and
Classification Activation Score (CAS).

\subsection{Self-supervised Framework}

As the base model training enables the model to perform snippet-level action
classes, we leverage the self-supervised method to adapt the model's output by
extracting contextual information. Our model is based on a mean teacher
framework~\cite{meanteacher,harmoniousteacher,consistentteacher}. As snippets
with high attention perform relevantly high classification accuracy
(\cref{problem}), our pipeline aims to extract these snippets to refine the
output of ambiguous snippets. Unlike the previous teacher-student structure that
performs different augmentations on the teacher and student models, our pipeline
targets refining the output of the teacher model to guide the training of the
student model.

In our framework, we initialize both the teacher model and student model with
the first-stage model in our method. We then convert the problem to a
self-supervised setting. Specifically, we separately predict
$\boldsymbol{\varphi}_T$, $\boldsymbol{\Psi}_T$ using the teacher model, and
$\boldsymbol{\varphi}_S$, $\boldsymbol{\Psi}_S$ using the student model for an
input video. $\boldsymbol{\varphi}$ is the attention and $\boldsymbol{\Psi}$ is
the classification activation score. Blurs are added to the input feature of the
student model as a weak augmentation to avoid overfitting. The attention
$\boldsymbol{\varphi}_S$ of the teacher model is refined in the Temporal
Refinement Block to obtain $\boldsymbol{\hat{\varphi}}_S$. Finally, we design
the Alignment Module to align the outputs of the student and teacher model.

\subsection{Temporal Refinement}
Due to the different action scales across distributions (\cref{problem}), a curated module is needed to enable adaptive learning.
Addressing this problem, we leverage curriculum learning, which iteratively generates pseudo-labels using salient snippets to guide their ambiguous counterparts.
To further improve performance, we leverage contextual information to refine the activation score of each snippet.

\noindent \textbf{Salience sampling.}
The relationship between the center snippet and its context strongly affects the model's preference for the scale of output proposals.
To leverage contextual information to refine the output of snippets, we first conduct salience sampling to extract the representative contextual snippets.
Specifically, we set a receptive field in a fixed size $\eta$ and the refinement factor $\alpha$ in this module.
For the attention $\boldsymbol{\varphi}_T$, we take a Salience Sampling strategy to sample the snippets with the highest attention score in the context. Our analysis in~\cref{problemlabel} shows that snippets with high attention scores exhibit high classification accuracy in both \SmD and \CrD testing.
Thus, we hypothesize that those snippets could be utilized as the salient snippets that could represent the actions with strong confidence. First, we extract the context of $\boldsymbol{\varphi}_T(n)$ with the left $\eta$ snippets denoted as $\boldsymbol{\varphi}_T(n-\eta:n-1) \in \mathbb{R}^{\eta \times 1}$, and the right $\eta$ snippets denoted as $\boldsymbol{\varphi}_T(n+1:n+\eta) \in \mathbb{R}^{\eta \times 1}$. Then we sample the most salient snippets separately on the left and right with:
\begin{equation}
\begin{aligned}
  & \boldsymbol{\varphi}_T^l(n) \leftarrow \max\big(\boldsymbol{\varphi}_T(n-\eta:n-1)\big),\\ & \boldsymbol{\varphi}_T^r(n) \leftarrow \max \big(\boldsymbol{\varphi}_T(n+1:n+\eta) \big),
  \label{equ:sampling}
\end{aligned}
\end{equation}
where $\boldsymbol{\varphi}_T^l(n) \in \mathbb{R}$ indicates the most salient snippet in the left context, and $\boldsymbol{\varphi}_T^r(n)$ for the right.
\begin{table*}[t]
    \setlength{\belowcaptionskip}{-0.2cm}
    \centering 
    \resizebox{0.9\linewidth}{!}{
    \begin{tabular}{l | c c c c c c c c | c c c c} 
        \toprule 
        \multirow{2}{*}{Methods} & \multicolumn{8}{c|}{ActivityNet to THUMOS14} & \multicolumn{4}{c}{THUMOS14 to ActivityNet} \\
        \cmidrule(lr){2-9} \cmidrule(lr){10-13} & 0.1 & 0.2 & 0.3 & 0.4 & 0.5 & 0.6 & 0.7 & Avg. & 0.5 & 0.75 & 0.95 & Avg.\\
        \midrule
        CO2-NET (\SmD) & 80.0 & 75.8 & 67.9 & 60.4 & 49.1 & 35.6 & 20.0 & 55.5 & 30.5 & 14.9 & 1.3 & 15.8 \\
        \midrule[0.1mm]
        CO2-NET & 51.4& 40.5 & 30.5 & 22.7 & 14.3 &7.3 & 3.2 & 24.3 & 9.5 & 2.8 & 0.1 & 3.5  \\
        CO2-NET + PGD & 59.5 & 45.4 & 37.3 & 29.3 & 20.5 & 12.8 & 5.4 & 30.0 & 12.3 & 3.3 & 0.1 & 4.5 \\
        CO2-NET + IRM & 53.8 & 43.6 & 33.0 & 23.9 & 14.9 & 7.5 & 3.2 & 25.2 & 15.3 & 4.0 & 0.3 & 5.4 \\
        CO2-NET + FPN & 65.1 & 61.6 & 56.0 & 47.4 & 37.9&24.7&15.0 & 44.0 & 24.0 & 8.0 & 0.6 & 10.1\\
        CO2-NET + SNIP & 66.0 &62.5 & 56.5& 47.4& 37.7 & 26.0 & 15.9 & 44.5 & 25.4 & 9.3 & 1.0 & 12.1 \\
        \midrule
        \textbf{Ours (CO2-NET)} & \textbf{73.5}& \textbf{71.6}& \textbf{67.2}& \textbf{59.4}& \textbf{48.0}& \textbf{34.0}& \textbf{19.8}& \textbf{53.4}& \textbf{28.7} & \textbf{12.8} & \textbf{1.1} & \textbf{14.4} \\
        \toprule
        DELU (\SmD) & 81.0 & 77.9 & 71.1 & 64.2 & 53.6 & 37.6 & 22.0 & 58.2 & 31.6 & 15.4 & 1.2 & 16.2 \\
        \midrule[0.1mm]
        DELU & 50.2 & 39.4 & 29.1 & 20.9 & 13.4 & 7.3 & 3.2 & 23.4 & 10.5 & 2.9 & 0.1 & 3.9 \\
        DELU + PGD & 59.7 & 46.9 & 38.5 & 29.5 & 20.7 & 12.3 & 5.2 & 30.4 & 11.8 & 3.2 & 0.2 & 4.5 \\ 
        DELU + IRM & 52.8 & 42.6 & 32.2 & 22.4 & 14.4 & 7.2 & 3.4 & 25.0 & 13.0 & 4.4 & 0.5 & 5.5 \\
        DELU + FPN & 65.1 & 61.6 & 55.0 & 46.4 & 36.7 & 26.3 & 14.7 & 44.0 & 25.2 & 8.3 & 0.3 & 10.5 \\
        DELU + SNIP & 66.5 & 58.2 & 54.0 & 46.9 & 39.0 & 27.5 & 16.5 & 43.3 & 30.5 & 12.9 & 0.8 & 14.5\\
        \midrule
        \textbf{Ours (DELU)} & \textbf{73.6} &\textbf{71.2} & \textbf{66.7} & \textbf{59.8} & \textbf{49.6} & \textbf{34.6} & \textbf{19.9} & \textbf{53.6} & \textbf{30.9} & \textbf{13.4} & \textbf{1.0} & \textbf{15.4} \\
        \toprule
        DDG-Net (\SmD) & 81.2 & 78.1 & 69.2 & 61.7 & 49.8 & 33.7 & 19.8 & 56.2 & 30.7 & 14.5 & 1.4 & 15.9 \\
        \midrule[0.1mm]
        DDG-Net & 47.3 & 36.7 & 26.1 & 18.0 & 11.8 & 5.8 & 2.2 & 21.1 & 9.6 & 2.5 & 0.2 & 3.7 \\
        DDG-NET + PGD & 54.5 & 40.8 & 36.5 & 29.5 & 22.2 & 13.6 & 6.6 & 29.1& 10.4 & 3.0 & 0.1 & 4.0 \\
        DDG-NET + IRM & 48.9 & 38.4 & 27.1 & 19.0 & 12.3 & 6.6 & 2.8 & 23.3 & 13.5 & 3.9 & 0.5 & 5.2 \\
        DDG-NET + FPN  & 62.2 & 58.2 & 51.4 & 43.8 & 34.2 & 22.9 & 13.2 & 40.8 & 25.0 & 8.8 & 0.5 & 11.3 \\
        DDG-NET + SNIP & 60.6 & 58.2 &  54.0 & 46.9 & 39.0 & 27.7 & 16.5 & 43.2 & 30.3 & 13.1 & 0.7 & 14.3 \\
        \midrule
        \textbf{Ours (DDG-Net)} &\textbf{71.8}&\textbf{69.3}& \textbf{65.3} & \textbf{58.7} & \textbf{50.2} & \textbf{36.4} & \textbf{22.8} & \textbf{53.5} & \textbf{30.3} & \textbf{12.9} & \textbf{1.2} & \textbf{15.1} \\
        \bottomrule 
    \end{tabular}
    }
    \caption{\CrD performance comparison on THUMOS14 and ActivityNet1.2. Note that the results for CO2-Net (\SmD), DELU (\SmD) and DDG-Net (\SmD) are on the sharing classes only, and only for reference.}
    \label{oodresults} 
\end{table*}

\noindent \textbf{Attention refinement.} To ensure the ranking coherency of the sampled snippets, a refinement rule was implemented, as illustrated in~\cref{pipeline}.
To prevent the scores of snippets with lower salience from surpassing those with higher scores, we established a refinement rule, which contains three different conditions. In condition (a), refinement was performed directly based on $\boldsymbol{\varphi}_T^l(n)$. However, in condition (b), refinement based on $\boldsymbol{\varphi}_t^l$ or $\boldsymbol{\varphi}_t^r$ would harm accuracy by guiding the refinement with more ambiguous snippets. In condition (c), the operation was not performed to maintain the geometry property. Therefore, we represent the rule with the following form:
\begin{equation}
  \begin{aligned}
    \boldsymbol{\hat{\varphi}}_T(n) =  \begin{cases} \alpha \boldsymbol{\varphi}_T(n) + (1-\alpha) \boldsymbol{\varphi}^{ref}_T(n) & \boldsymbol{\varphi}_T(n)<\boldsymbol{\varphi}^{ref}_T(n) \\
    \boldsymbol{\varphi}_T(n) & \text{otherwise},
    \end{cases} \\
    \label{equ:refinement}
  \end{aligned}
\end{equation}
where $\boldsymbol{\varphi}_T^{ref}(n) = \min(\boldsymbol{\varphi}_T^l(n), \boldsymbol{\varphi}_T^r(n))$ represents the reference attention score for refinement, and the $\min(\cdot)$ operation guarantees the ranking coherency. $\alpha$ is a hyper-parameter to adjust the scale.

After refinement, $\boldsymbol{\hat{\varphi}}_T(n)$ is better adapted to the target dataset and could be used as a guidance for CAS refinement.

\subsection{Alignment Module}

The alignment module is crafted to enhance the Class Activation Sequence (CAS) by using attention values.
It achieves this by aligning the attention outputs from both the student and teacher models, which are obtained post-processing – specifically, after generating initial outputs from the student model and further refining them through the teacher model.

Specifically, we first align the attention output between the student and teacher models.
Next, we calibrate the attention and CAS of the student model outputs. Finally, we align the output of the CAS score between the two models. To guide the student model to adapt to the new scale of the teacher model, we apply an MSE loss:
\begin{equation}
  L_{att} = \frac{1}{N}\sum_{n=1}^{N}\big(\boldsymbol{\hat{\varphi}}_T(n) - \boldsymbol{\varphi}_S(n) \big)^2
  \label{eq:att}
\end{equation}
where $\boldsymbol{\hat{\varphi}}_T(n)$ denotes the attention score of the $n$-th snippet from the is the refined attention from the teacher model from~\cref{equ:refinement}, and $\boldsymbol{\varphi}_S(n)$ is its counterpart from the student model.
We use the output $\boldsymbol{\varphi}_S$ directly predicted by the student to align the refined attention score $\boldsymbol{\hat{\varphi}}$ from the teacher.

Then, we align the CAS of the teacher and the student model with a KL-divergence loss $L_{cas}$:
\begin{equation}
\begin{aligned}
L_{cas}=\frac{1}{N}\sum_{n=1}^N\sum_{c=1}^{C_I+1}\boldsymbol{\Psi}_T(n,c)\log \frac{\boldsymbol{\Psi}_T(n,c)}{\boldsymbol{\Psi}_S(n,c)},
\label{equ:cas}
\end{aligned}
\end{equation}

An ideal situation is that the predicted background probability to be complementary with the attention.
Thus, we propose a calibration loss $L_{cal}$ for the alignment:
\begin{equation}
\begin{aligned}
    L_{cal} = & \frac{1}{N} \sum_{t=1}^{T} [-\boldsymbol{\varphi}_S(n) \log(\boldsymbol{\Psi}_S(n,C_I+1)) \\ & - (1-\boldsymbol{\varphi}_S(n) \log(1-\boldsymbol{\Psi}_S(n,C_I+1))],
\label{equ:sal}
\end{aligned}
\end{equation}
where $\boldsymbol{\Psi}_S(n,C_I+1))$ represents the predicted probability for the background. This loss enables alignment between attention and CAS, and makes both consistent to discern foreground and background.

\subsection{Training and Inference}

\textbf{Training.} By combining all optimization objectives introduced above, we obtain the overall loss function:
\begin{equation}
  L=\lambda_1 L_{att}+\lambda_2 L_{cas}+\lambda_3 L_{cal},
  \label{equ:finalloss}
\end{equation}
where $\lambda_1$, $\lambda_2$, and $\lambda_3$ are weight hyper-parameters.
The overall loss $L$ is only applied to the student model for parameter update, while the teacher model is updated with an exponential moving average (EMA).

\begin{table}[t]
    \setlength{\belowcaptionskip}{-0.2cm}
    \centering
    \begin{minipage}[t]{0.5\textwidth}
    \resizebox{\linewidth}{!}{
    \begin{tabular}{l | c c c c | c c c c} 
        \toprule 
        \multirow{2}{*}{Methods} & \multicolumn{4}{c|}{THUMOS14 to HACS} & \multicolumn{4}{c}{ActivityNet to HACS} \\
        \cmidrule(lr){2-5} \cmidrule(lr){6-9} & 0.5 & 0.75 & 0.95 & Avg. & 0.5 & 0.75 & 0.95 & Avg.\\
        \midrule
        CO2-Net & 21.5 & 7.4 & 0.3 & 9.3 & 17.6 & 6.2 & 0.9 & 8.7 \\
        + PGD & 24.9 & 8.4 & 0.4 & 9.9 & 17.9 & 6.2 & 0.8 & 8.8 \\
        + IRM & 18.4 & 6.1 & 0.3 & 9.5 & 23.3 & 6.9 & 1.0 & 9.3 \\
        + FPN & 30.7 & 11.0 & 0.8 & 13.9 & 25.3 & 8.5& 0.8 & 11.0 \\
        + SNIP & 30.4 & 12.8 & 0.9 & 14.2 & 26.5 & 9.5 & 0.8 & 11.7 \\
        \midrule
        \textbf{Ours (CO2-Net)} & \textbf{29.0} & \textbf{8.0} & \textbf{0.8} & \textbf{14.7} & \textbf{26.6} & \textbf{9.4} & \textbf{1.0} & \textbf{12.3} \\
        \toprule
        DELU & 24.1 & 8.0 & 0.4 & 10.0 & 17.0 & 6.6 & 0.8 & 8.0 \\
        + PGD & 26.8 & 8.9 & 0.5 & 10.6 & 18.2 & 6.8 & 0.7 & 8.4 \\
        + IRM & 24.4 & 8.2 & 0.4 & 10.1 & 24.0 & 6.8 & 0.6 & 9.7  \\
        + FPN & 27.4 & 8.7 & 0.7 & 12.3 & 27.4 & 8.7 & 0.7 & 11.5\\
        + SNIP & 31.3 & 11.6 & 0.9 & 13.8 & 27.6 &10.2 & 1.0& 12.5\\
        \midrule
        \textbf{Ours (DELU)} & \textbf{36.4} & \textbf{12.3} & \textbf{0.8} & \textbf{15.8} &  \textbf{30.1} & \textbf{11.0} & \textbf{1.0} & \textbf{13.0}\\
        \toprule
        DDG-Net& 21.0 & 7.3 & 0.3 & 8.8 & 16.4 & 6.4 & 1.0 & 7.9 \\
        + PGD & 22.4 & 7.8 & 0.4 & 9.5 & 17.0 & 6.8 & 1.0 & 8.1 \\
        + IRM & 21.5 & 7.5 & 0.4 & 9.0  & 22.8 & 7.0 & 0.8 & 9.2 \\
        + FPN & 28.0 & 10.3 & 0.6 & 12.1 & 27.8 & 8.7 & 0.8 & 11.8 \\
        + SNIP & 30.5 & 10.0 & 1.1 & 13.4 & 25.6 & 9.6 & 1.2 & 11.9\\
        \midrule
        \textbf{Ours (DDG-Net)} & \textbf{33.7} & \textbf{12.6} & \textbf{0.5} & \textbf{15.4} & \textbf{29.2} & \textbf{10.4} & \textbf{1.1} & \textbf{12.7}\\
        \bottomrule 
    \end{tabular}
    }
    \caption{\CrD~performance comparison on the HACS dataset.} 
    \label{hacsresults}
    \end{minipage}
    \hfill
    \begin{minipage}[t]{0.45\textwidth}
    \resizebox{\linewidth}{!}{
    \begin{tabular}{l |c | c c c | c c c c} 
        \toprule 
        \multirow{2}{*}{Exp} & \multirow{2}{*}{Teacher} & \multicolumn{3}{c|}{Loss} & \multicolumn{4}{c}{ActivityNet} \\
        \cmidrule(lr){3-5} \cmidrule(lr){6-9} & \multirow{1}{*}{\centering} & $L_{att}$ &$L_{cas}$ & $L_{cal}$ &  0.5 & 0.75 & 0.95 & Avg.\\
        \midrule
        1 & \ding{55} & \ding{55}& \ding{55}& \ding{55} & 10.5 & 2.9 & 0.1 & 3.9 \\
        \midrule
        2 & \ding{55} & \ding{55}& \checkmark & \checkmark & 12.4 & 3.9 & 0.1 & 4.8 \\
        3 & \ding{55}&\checkmark &\ding{55} &\ding{55} & 23.4 & 9.8 & 0.8 & 10.9  \\
        4 & \ding{55}&\checkmark &\checkmark &\ding{55} & 23.5 & 9.8 & 0.8 & 10.9  \\
        5 & \ding{55} & \checkmark & \ding{55} & \checkmark & 23.6 & 9.9 & 0.8 & 11.0 \\
        6 & \ding{55} & \checkmark & \checkmark & \checkmark & 23.7 & 9.9 & 0.9 & 11.1 \\
        \midrule
        7 & \checkmark&\ding{55} &\checkmark &\checkmark & 18.3 & 5.0 & 0.5 & 7.7  \\
        8 & \checkmark& \checkmark & \ding{55} & \ding{55} & 30.4 & 12.6 & 0.9 & 15.0 \\
        9 & \checkmark& \checkmark & \checkmark & \ding{55} & 30.8 & 12.8 & \textbf{1.0} & 15.1 \\
        10 & \checkmark & \checkmark & \ding{55} & \checkmark & 30.8 & 12.9 & \textbf{1.0} & 15.3 \\
        11 & \checkmark & \checkmark &\checkmark  & \checkmark & \textbf{30.9} & \textbf{13.4} & \textbf{1.0} & \textbf{15.4} \\
        \bottomrule 
    \end{tabular}
    }
    \caption{Ablation study on the teacher-student framework, and the three loss terms.}
    \label{tab:ablationt2a}
    \end{minipage}
\end{table}

\noindent \textbf{Inference.} After the two-stage training, we use the teacher model for inference.
We predict the action categories for each video using \cref{equ:base_sup}.
To obtain the closed-set video-level classification score, we first apply attention to filter the top-k scores for each class, and then average them. We use the score to generate candidate action proposals by applying multiple thresholds on the attention sequence for each class separately. Finally, we apply soft-NMS~\cite{softnms} to these proposals to obtain the final predictions.

\section{Experiments}
\label{sec:experiments}

\textbf{Datasets.}
We use three datasets with different distributions: THUMOS14~\cite{thumos}, ActivityNet1.2~\cite{activitynet}, and HACS~\cite{hacs}. The THUMOS14 dataset consists of 200 validation videos and 213 testing videos, spanning 20 categories, with each video containing an average of 15.4 action instances. ActivityNet1.2 contains 9,682 videos across 100 categories. Each video has an average duration of 51.4 seconds and contains approximately 1.5 action instances. HACS contains 5,981 testing videos, spanning 200 categories, with each video having an average length of 40.6 seconds and approximately 1.8 action instances. Among these three datasets, 8 action classes are semantically equivalent. For cross-distribution (\CrD) testing, we selectively sample from the test sets of these datasets. Specifically, we chose a subset of 83 samples from THUMOS, 193 samples from ActivityNet1.2, and 210 samples from HACS.

\noindent \textbf{Metrics.}  We evaluate the TAL performance with mean Average Precision (mAP) at different Intersections over Union (IoU) thresholds. The IoU thresholds for THUMOS14 are set at [0.1:0.1:0.7], for ActivityNet1.2 at [0.5:0.05:0.95], and for HACS at [0.5:0.05:0.95].

\noindent \textbf{Implementation Details.} In our experiments, we use DELU~\cite{delu} and DDG-Net~\cite{ddgnet} and CO2-Net~\cite{co2net} as our baselines. To extract both RGB and optical flow features, we use Kinetics~\cite{kinetics} pre-trained I3D~\cite{i3d}.
The extracted features are in the dimension of 1024. For the THUMOS14 and ActivityNet1.2 datasets, we extract one snippet feature for every 16 frames, with a video fps of 25, which is consistent with the setting in CO2-Net~\cite{co2net}. For the HACS dataset, we extract each snippet feature every 16 frames, following the original video fps. We set $\eta$ to 3, $\alpha$ to 1.4 for THUMOS14 to Activitynet1.2 and HACS, and $\eta$ to 22, $\alpha$ to 0.1 for ActivityNet to THUMOS14 and HACS. We set the total epochs to 50 for THUMOS14 to ActivityNet1.2 and HACS, and 200 for ActivityNet1.2 to THUMOS14. $\lambda_1$ and $\lambda_2$ are set as 1.0, and $\lambda_3$ are set as 0.1. The learning rate is 3e-5, and the dropout ratio for the student model is 0.1. The batch size is set to 30, while the EMA update parameter is set to 0.9.

\subsection{Comparison with State-of-the-art Methods}

We compare our proposed method with baselines based on CO2-Net, DELU, and DDG-Net. To the best of our knowledge, there is no previous work on WTAL that directly focuses on generalizability. Therefore, we compare our method with widely used strategies in image classification, object detection, and other downstream tasks.
Following the GTAL setting, the models could only be trained and tuned on the training set of the SmD dataset.
The first two are general OODG methods, and the last two are common scale variance handlers.
(1) \textbf{PGD}~\cite{pgdattack}: Projected Gradient Descent is an iterative gradient-based method that aims to generate adversarial examples by iteratively perturbing the input to help the model better adapt to an out-of-distribution dataset.
(2) \textbf{IRM}~\cite{irm}: Invariant Risk Minimization is a framework that aims to improve the generalization of models to out-of-distribution data by learning invariant features that are robust to distribution shifts.
(3) \textbf{FPN}~\cite{fpn}: Feature Pyramid Network is an architecture widely used in various downstream tasks to improve performance on multi-scale instances.
(4) \textbf{SNIP}~\cite{snip}: a strategy to deal with scale variance during training. \footnote{The scale for the CrD test is specially tuned for our version of FPN and SNIP.}

\noindent \textbf{Comparison of CrD results.}
\cref{oodresults} compares the \CrD~localization results on the THUMOS14 and ActivityNet1.2 dataset.
The results show that our proposed method consistently outperforms the baselines by a large margin and even approaches \SmD~results.
For \CrD~testing, we notice that the three models have consistent results.
For PGD and IRM, the two general OODG methods that focus on avoiding the model to overfit and better robustness to blurs, can bring limited improvement.
As for FPN and SNIP, the two methods that focus on dealing with scale variance gain much better performance, proving our analysis.

\cref{hacsresults} shows the \CrD~localization results on the HACS dataset.
Our findings on HACS closely resemble those on THUMOS14 and ActivityNet1.2. While PGD and IRM show limited improvement, FPN and SNIP deliver superior performance. With the HACS dataset falling between the scales of THUMOS14 and ActivityNet1.2, as illustrated in \cref{duration}, our model excels in both scaling up and scaling down scenarios, leading to significant improvements.

\begin{figure}[t]
  \centering
       \centering
       \includegraphics[width=0.8\linewidth]{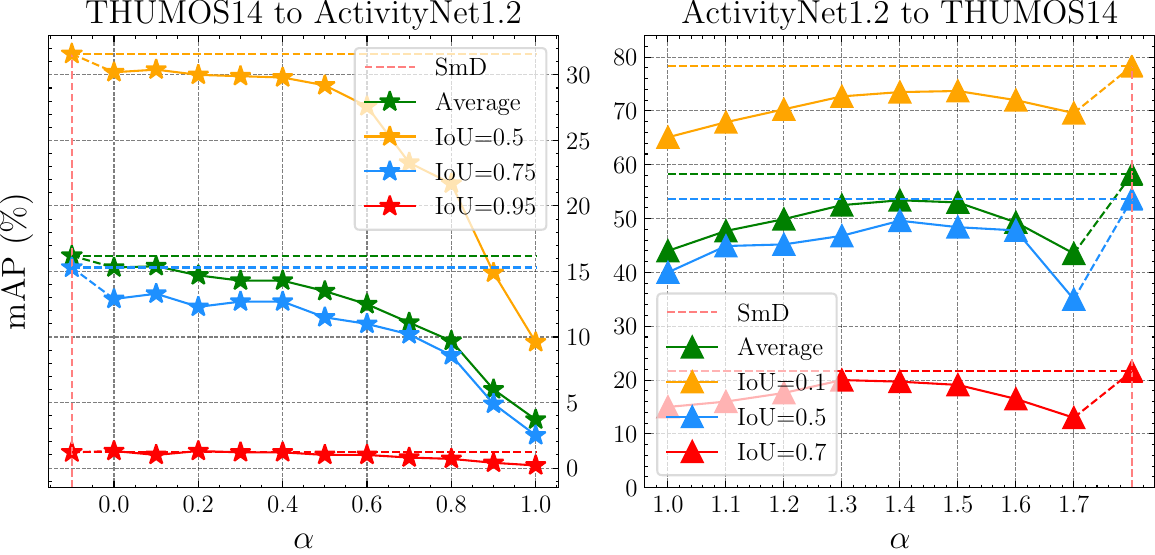}
       \caption{Ablation study on the refinement parameter $\alpha$. 
       Left: THUMOS14 to ActivityNet1.2. Right: ActivityNet1.2 to THUMOS14.}
       \label{alpha}
\end{figure}

\subsection{Ablation Study}

In this section, We perform experiments on THUMOS14 to ActivityNet1.2 and use DELU as our base model.

\noindent\textbf{Model architecture.}
\cref{tab:ablationt2a} details the effects of incrementally integrating each component of our model. We conducted an ablation study focusing on the teacher-student structure and the three losses within the alignment module.
It is notable that in the absence of the teacher-student structure, we simply freeze the teacher model for evaluation with the student model.
The comparison between Exp.~2-6 (without teacher-student structure) and Exp.~7-11 (with teacher-student structure) reveals that the latter consistently yields significantly better performance.
This suggests that the teacher-student framework improves the robustness of the model against visual gap and lays a strong foundation for enhancing the effectiveness of the other three components.
Notably, even with the sole application of $L_{att}$ in Exp.~3 and Exp.~8, we observe CrD performance nearing that of the complete pipeline, highlighting the predominant role of $L_{att}$.
Regarding the other two losses, $L_{cas}$ and $L_{cal}$, as shown in Exp.~2, 4, 5, 7, 9, and 10, their individual impact is less pronounced, as both primarily function to calibrate the CAS score output. However, these losses do contribute to improved performance when combined with $L_{att}$.

\noindent \textbf{Scale parameters.}
We analyze the effectiveness of the refinement scale parameter $\alpha$, which enables the model to adjust to different scales. We presume that it is feasible to estimate the scale relationship between SmD and CrD.
Specifically, we tested various $\alpha$ values using DELU for two scenarios: scaling up from THUMOS14 to ActivityNet1.2 and scaling down from ActivityNet1.2 to THUMOS14. The results of these tests are presented in \cref{alpha}.
For the scaling up from THUMOS14 to ActivityNet1.2, we experimented with $\alpha$ values ranging from 0.0 to 1.0. This range is to enlarge the scale for ActivityNet1.2, which is a coarser dataset.
Conversely, for scaling down from ActivityNet1.2 to THUMOS14, we tested $\alpha$ values from 1.0 to 1.7 for a finer scale adaptation. It's noteworthy that values of $\alpha$ exceeding 1.6 tend to result in performance degradation. We observed stable performance within the intervals of [0.0 to 0.5] and [1.1 to 1.5] for the respective scaling up and down scenarios. Overall, our method shows a notable improvement in GTAL and maintains stability over a wide range of $\alpha$ values.

\section{Conclusion}

To improve the generalizability of Temporal Action Localization (TAL) methods, we propose the task of Generalizable Temporal Action Localization (GTAL).
Addressing the problem, we propose a \ModelFullName (\ModelName) framework.
Specifically, with teacher-student structure, \ModelName~features a refinement module and an alignment module. The former iteratively refines the teacher model output to CrD distribution, and the latter aligns the teacher and student outputs with different priors, promoting precise CrD localization.
Our comprehensive experiments demonstrate that our method achieves state-of-the-art performance on CrD results. Notably, the CrD results of \ModelName can even approach the SmD results from existing methods.

\noindent\textbf{Limtations.}
(1) Though STAT could perform high \CrD results, achieving consistently high results across both \SmD and \CrD evaluations without tuning remains a significant challenge.
(2) We observed that action annotations significantly differ across scenarios. Currently, our methods are limited to class-agnostic scale adaptation, leaving room for future improvements in class-aware annotation discrepancies.
(3) The GTAL task is premised on the high cost of segment-level annotations, restricting the pre-trained model's learning of scale variance. Future research may investigate enabling the pre-trained model to effectively use the segment-level annotations in the \CrD dataset for improved learning.

\section*{Acknowledgement}
This work is supported in part by the Beijing Natural Science Foundation (No.~4244082) and the Defense Advanced Research Projects Agency (DARPA) under Contract No.~HR001120C0124. Any opinions, findings and conclusions or recommendations expressed in this material are those of the author(s) and do not necessarily reflect the views of the Defense Advanced Research Projects Agency (DARPA).
\newpage
\bibliographystyle{splncs04}
\bibliography{main}
\end{document}